\documentclass[10pt,twocolumn,letterpaper]{article}

\usepackage[pagenumbers]{wacv} %
\usepackage{graphicx}
\usepackage{tikz}
\usepackage{pgfplots}
\usepgfplotslibrary{groupplots}
\pgfplotsset{compat=1.18}
\usetikzlibrary{calc, positioning, arrows.meta, shapes.geometric, fadings, shadows.blur}
\usetikzlibrary{fadings}
\tikzfading[name=fade out,
  inner color=transparent!0,
  outer color=transparent!100
]
\usepackage[table]{xcolor}
\newcommand{\best}{\cellcolor{tablered}}
\newcommand{\sbest}{\cellcolor{orange}}
\newcommand{\tbest}{\cellcolor{yellow}}

\definecolor{yellow}{rgb}{1, 1, 0.7}
\definecolor{orange}{rgb}{1, 0.85, 0.7}
\definecolor{tablered}{rgb}{1, 0.7, 0.7}

\definecolor{wacvblue}{rgb}{0.21,0.49,0.74}
\usepackage[pagebackref,breaklinks,colorlinks,allcolors=wacvblue]{hyperref}

\def\wacvPaperID{159} %
\def\confName{WACV}
\def\confYear{2027}
\def\our{TOM-GS}

\title{\our{}: Editable Video Representation via Temporal Opacity Modulation of Static 3D Gaussians}

\author{
Marek Lisowski$^{*1}$
\qquad
Łukasz Smoliński$^{*1}$
\qquad
Kornel Howil$^{2,3}$
\qquad
Piotr Biliński$^1$
\qquad\\
Marcin Mazur$^2$
\qquad
Przemysław Spurek$^{2,3}$
\vspace*{0.2cm}
\\
$^1$University of Warsaw, Faculty of Mathematics, Informatics, and Mechanics \qquad\\ $^2$Jagiellonian University, Faculty of Mathematics and Computer Science \qquad\\ $^3$IDEAS Research Institute \qquad\\ 
$^*$Equal contribution
}

\begin{document}

\maketitle

\begin{abstract}
While Implicit Neural Representations (INRs) and dynamic 3D Gaussian Splatting (3DGS) achieve impressive results in video processing, they often fall short of producing representations that are easily editable. Recent methods address this by introducing complex spatial deformations or folded distributions, which constrain optimization and reduce flexibility for downstream editing. In this paper, we introduce \our, an editable video representation that forgoes complex deformations in favor of regular 3D Gaussians equipped with a continuous temporal opacity formulation. By assigning a learnable temporal mean and scale to the opacity of each Gaussian, our model enables static 3D spatial components to fade smoothly in and out of the scene. Grounded by robust, off-the-shelf pose estimation, our approach maintains a static spatial geometry that naturally supports a wide range of manual and physics-based edits. \our{} outperforms prior editable video representations in visual fidelity, while its reliance on standard 3D Gaussians ensures seamless compatibility with established 3D editing tools.
\end{abstract}

\input{figures/teaser.tex}

\section{Introduction}

Decomposing video data into structured, continuous representations is a critical step toward advanced video processing and editing. While Implicit Neural Representations (INRs) can effectively encode the spatiotemporal signal of a video into the weights of a neural network, their black-box nature makes them notoriously difficult to edit or manipulate in a controlled manner. To overcome this, recent work has shifted toward explicit and semi-explicit representations. In particular, the 3D Gaussian Splatting (3DGS) framework~\citep{kerbl20233dgaussiansplattingrealtime} has proven a strong foundation for spatial representations, combining the rendering efficiency of rasterization with the flexibility of point-based geometry.

Adapting 3DGS to unconstrained videos, however, introduces the challenge of modeling nonlinear temporal dynamics and complex topological changes. Previous approaches, such as Splatter-a-Video~\cite{sun2024splattervideovideogaussian}, rely on canonical-space Gaussians coupled with deformation fields. While these support certain basic transformations, they struggle with structural editing over time. To enable realistic and comprehensive modifications, VeGaS~\cite{smolakdyzewska2026vegas} proposed a family of Folded-Gaussian distributions that model nonlinear dynamics through conditional Gaussian structures. This formulation showed that Gaussian-based video representations can support global operations such as multiplication, scaling, and spatial editing. While effective, this expressiveness stems from a specialized folded representation rather than from standard 3DGS primitives. As a result, adapting conventional 3DGS editing operations or geometry-based simulation pipelines requires reasoning through the folded parameterization.

To address these limitations, we propose \our, an editable video representation that preserves the spatial structure unchanged and models scene dynamics exclusively through temporal opacity, unlike VeGaS. This design trades a degree of representational expressiveness for a simpler, more directly editable 3D structure, and in doing so demonstrates that complex spatial deformations or folded distributions are unnecessary for capturing dynamic video content. Instead, \our{} relies entirely on regular 3D Gaussians and augments them with a \textit{temporal opacity modulation}. By assigning a learnable mean and scale to the opacity of each Gaussian along the temporal axis, our model allows static 3D spatial components to fade in and out of the scene. Because the underlying geometry remains standard 3DGS, it stays compatible with existing 3D editing tools, physics engines, and manual manipulation workflows, while still modeling moving objects, disocclusions, and transient effects.

A core contribution of our approach is a novel pipeline that leverages AnyCam~\cite{wimbauer2025anycamlearningrecovercamera} for robust feed-forward camera tracking, enabling fully automatic anchoring of our temporal representation in a consistent global space without any manual pose annotation. This allows our static Gaussians to be unambiguously grounded in a single world frame, a critical prerequisite for disentangling appearance from dynamics. Crucially, we further introduce a set of tailored training heuristics that, for the first time, address the unique optimization challenges posed by temporal opacity learning: we disable standard opacity resets to preserve learned temporal structures across training, and propose a novel inverse PSNR-based frame sampling strategy that adaptively concentrates representational capacity on the most dynamic and challenging segments of the video, yielding more faithful reconstructions where they matter most.

To sum up, the primary contributions of our work are as follows:

\begin{itemize}

\item We propose \our, a video representation built entirely from static 3D Gaussians augmented with a learnable temporal mean and scale for opacity, capturing non-rigid scene dynamics without complex spatial deformations.

\item We show that \our{} achieves state-of-the-art reconstruction quality among editable video representations on the DAVIS dataset~\citep{ponttuset20182017davischallengevideo}, demonstrating that high visual fidelity requires neither deformation fields nor folded distributions; we attribute this to optimization heuristics tailored to temporal opacities, namely disabling standard opacity resets and inverse PSNR-based frame sampling.

\item We demonstrate that retaining static spatial coordinates makes the representation directly compatible with standard 3D editing tools, enabling explicit manual modifications and physics-based simulations on video data.

\end{itemize}

\section{Related Work}

\paragraph{Video Representations and Editing.}
Early approaches to controllable video editing rely on two-dimensional layered representations or on deformation fields applied to a single canonical image. CoDeF~\cite{ouyang2024codefcontentdeformationfields}, for instance, maps every frame to a canonical image so that an edit applied once propagates through time. Such canonical formulations are effective for largely planar, slowly varying content, but degrade under large topological changes, disocclusions, and high-frequency motion, where no single canonical image can account for all observed appearance. Implicit Neural Representations (INRs), exemplified by the NeRV family~\cite{chen2021nerv, li2022nerv, chen2023hnerv}, instead encode the spatiotemporal signal of a video in the weights of a network, offering strong compression but little explicit structure, and their black-box nature makes localized, controlled manipulation difficult. These limitations motivate explicit, geometry-aware representations that expose primitives an editor can act on directly.

\paragraph{Dynamic 3D Gaussian Splatting.}
Before the rise of explicit splatting, dynamic novel-view synthesis was largely driven by neural radiance fields extended to time, either through deformation toward a canonical frame~\cite{pumarola2021d, park2021nerfies} or through higher-dimensional embeddings that absorb topological change~\cite{park2021hypernerf}. 3D Gaussian Splatting (3DGS)~\cite{kerbl20233dgaussiansplattingrealtime} has since become a widely adopted explicit alternative, and a substantial body of work extends it to dynamic content. One family attaches an explicit deformation field to a set of canonical Gaussians and predicts per-Gaussian motion at each timestep~\cite{yang2024deformable, wu20244d}; another lets the Gaussians themselves translate and rotate over time under local-rigidity constraints~\cite{luiten2024dynamic}. In the monocular video setting, Splatter-a-Video~\cite{sun2024splattervideovideogaussian} couples canonical Gaussians with learned 3D motions but supports only a restricted set of transformations, while VeGaS~\cite{smolakdyzewska2026vegas} introduces a family of Folded-Gaussian distributions that capture nonlinear dynamics through 2D conditional Gaussians and thereby enable a broader range of edits. These methods improve temporal expressiveness, but they entangle appearance with time-varying geometry: the resulting primitives no longer behave like a standard 3D asset and are awkward to manipulate with conventional tools or physics engines.

\paragraph{Editing and Simulation of Gaussian Scenes.}
A growing body of work manipulates 3DGS scenes directly. Instruction- and text-driven systems such as GaussianEditor~\cite{chen2024gaussianeditor} edit appearance and geometry by updating Gaussian attributes under guidance from 2D diffusion models, while physics-integrated formulations such as PhysGaussian~\cite{xie2024physgaussian} treat the Gaussians as material points and evolve them with a continuum-mechanics solver, producing physically plausible deformation and motion. A common premise underlies these pipelines: they operate on a static, well-formed 3DGS asset whose Gaussians represent a fixed geometry that the editing operator or solver then transforms. Representations whose primitives already translate, rotate, or fold over time violate this premise, forcing any such operator to reason through a learned deformation or folded parameterization rather than acting on the geometry directly.

\begin{figure}
    \centering
  \captionsetup[subfigure]{labelformat=empty, skip=1pt}

  \makebox[14pt]{\rotatebox{90}{\scriptsize Original}}%
  \begin{subfigure}[b]{0.23\linewidth}
    \includegraphics[width=\linewidth]{}
  \end{subfigure}\hfill
  \begin{subfigure}[b]{0.23\linewidth}
    \includegraphics[width=\linewidth]{}
  \end{subfigure}\hfill
  \begin{subfigure}[b]{0.23\linewidth}
    \includegraphics[width=\linewidth]{}
  \end{subfigure}\hfill
  \begin{subfigure}[b]{0.23\linewidth}
    \includegraphics[width=\linewidth]{}
  \end{subfigure}\hfill

  \makebox[14pt]{\rotatebox{90}{\scriptsize Physics}}%
    \begin{subfigure}[b]{0.23\linewidth}
    \includegraphics[width=\linewidth]{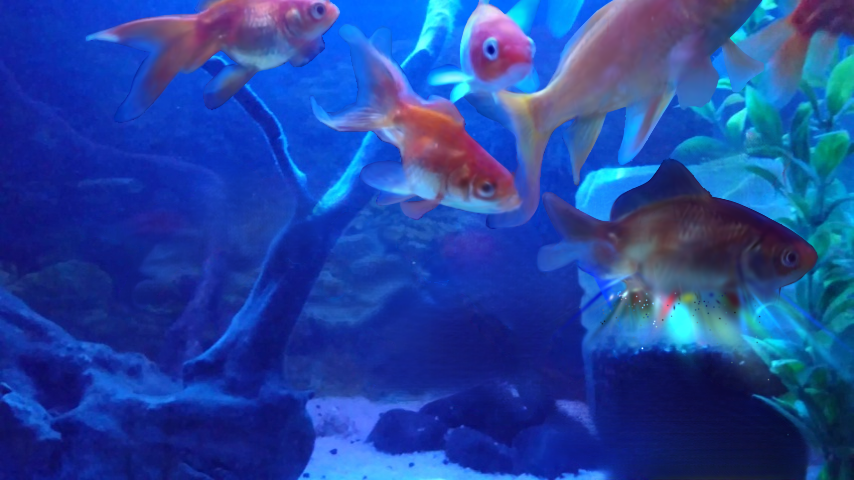}
  \end{subfigure}\hfill
  \begin{subfigure}[b]{0.23\linewidth}
    \includegraphics[width=\linewidth]{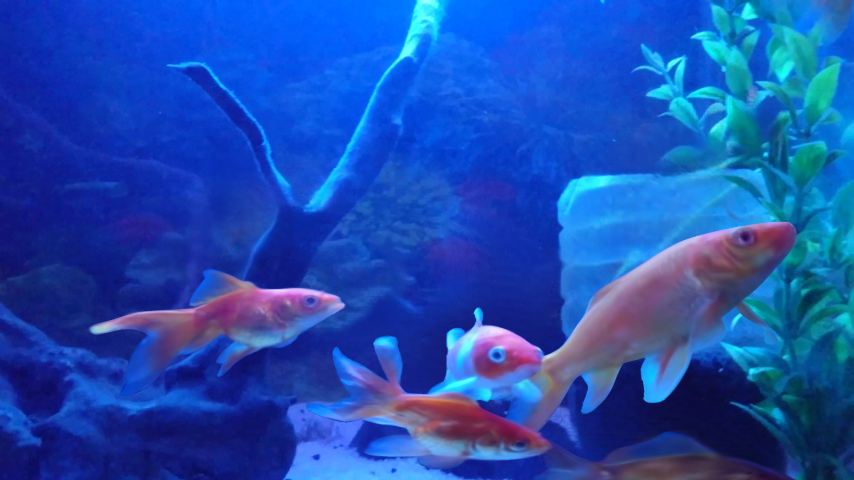}
  \end{subfigure}\hfill
  \begin{subfigure}[b]{0.23\linewidth}
    \includegraphics[width=\linewidth]{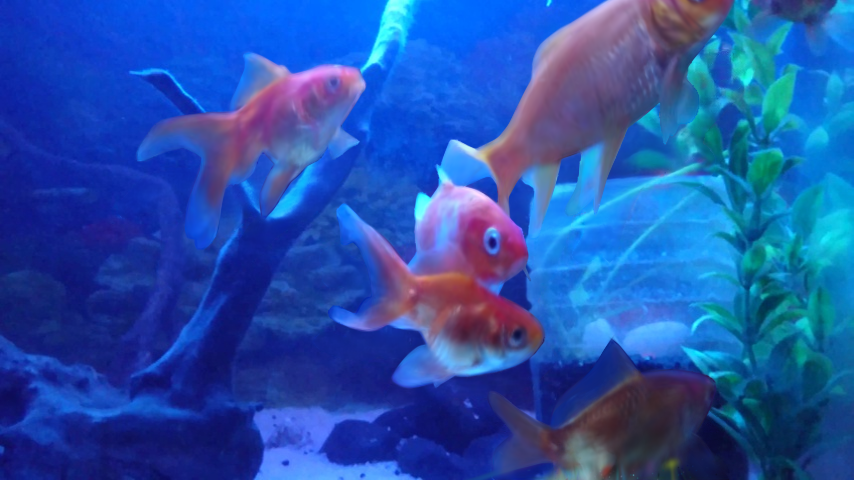}
  \end{subfigure}\hfill
  \begin{subfigure}[b]{0.23\linewidth}
    \includegraphics[width=\linewidth]{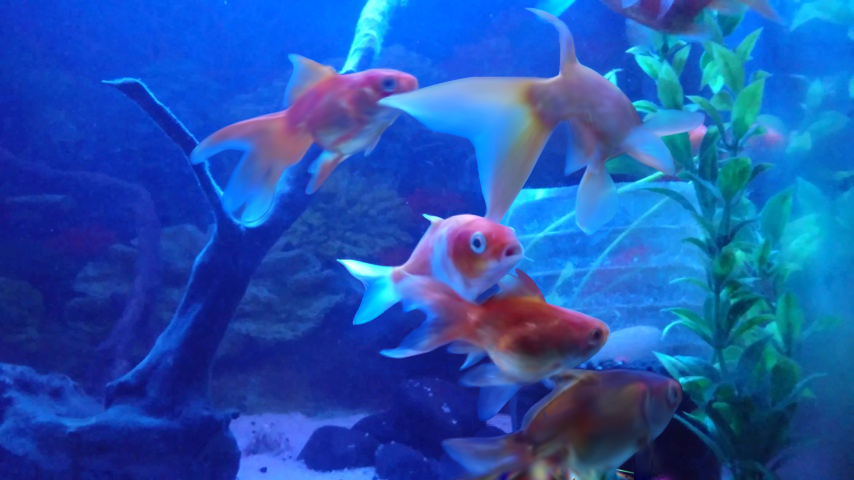}
  \end{subfigure}\hfill

  \makebox[14pt]{\rotatebox{90}{\scriptsize Original}}%
  \begin{subfigure}[b]{0.23\linewidth}
    \includegraphics[width=\linewidth]{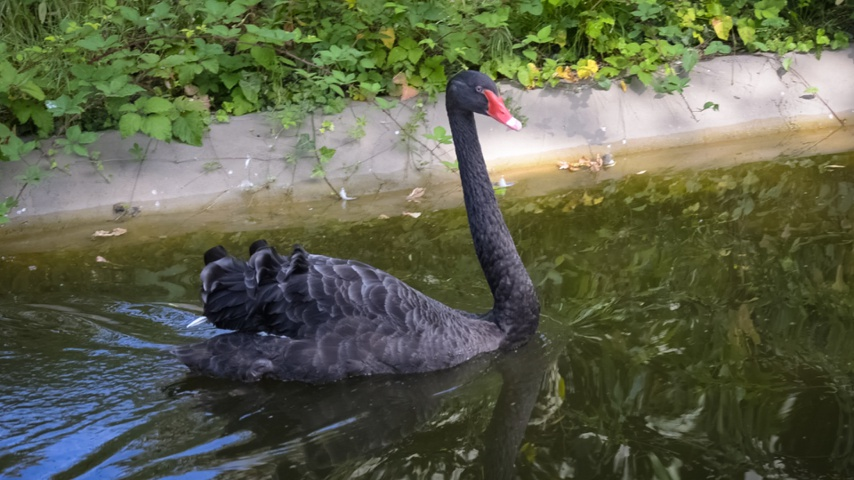}
  \end{subfigure}\hfill
  \begin{subfigure}[b]{0.23\linewidth}
    \includegraphics[width=\linewidth]{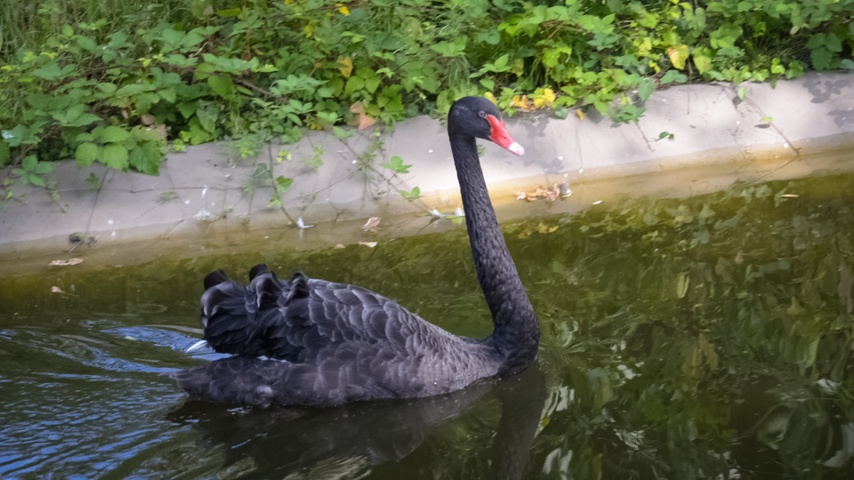}
  \end{subfigure}\hfill
  \begin{subfigure}[b]{0.23\linewidth}
    \includegraphics[width=\linewidth]{}
  \end{subfigure}\hfill
  \begin{subfigure}[b]{0.23\linewidth}
    \includegraphics[width=\linewidth]{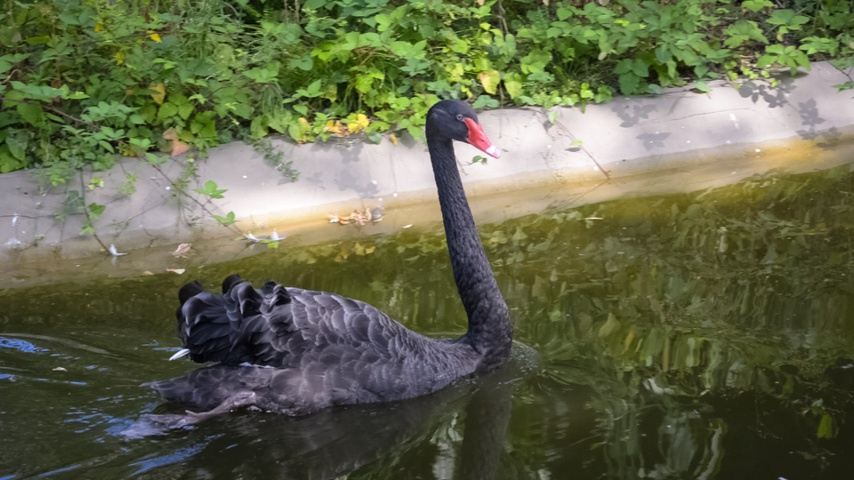}
  \end{subfigure}\hfill

  \makebox[14pt]{\rotatebox{90}{\scriptsize Physics}}%
    \begin{subfigure}[b]{0.23\linewidth}
    \includegraphics[width=\linewidth]{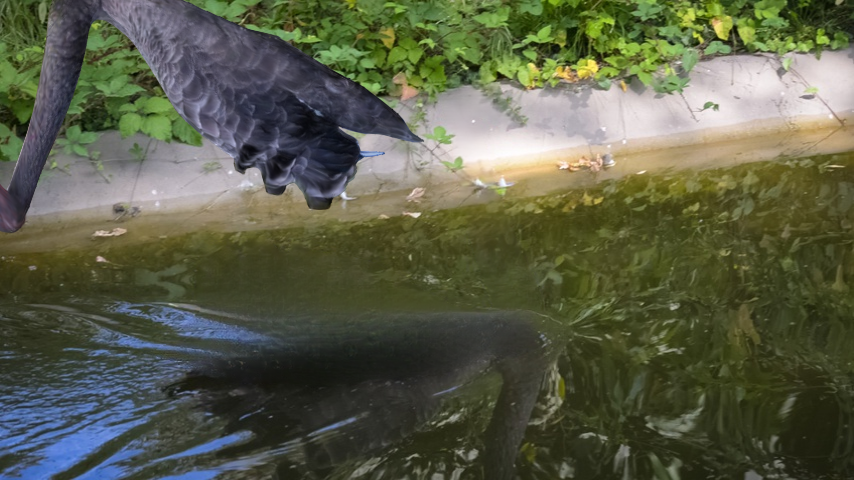}
  \end{subfigure}\hfill
  \begin{subfigure}[b]{0.23\linewidth}
    \includegraphics[width=\linewidth]{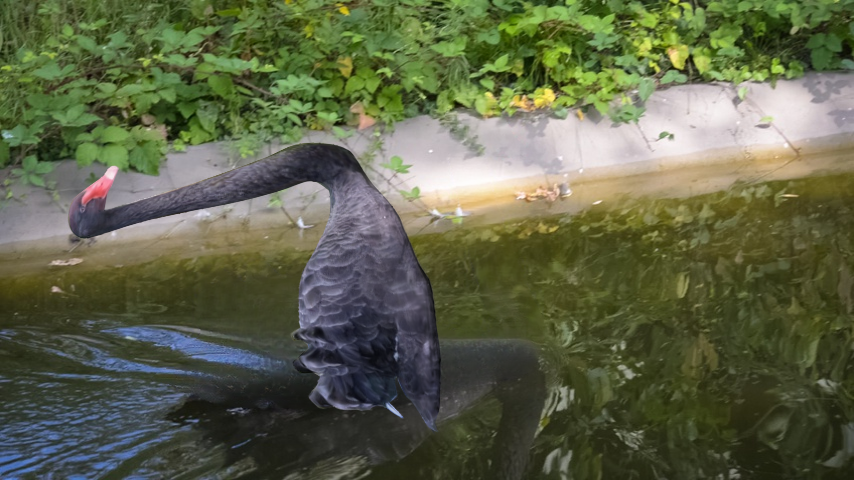}
  \end{subfigure}\hfill
  \begin{subfigure}[b]{0.23\linewidth}
    \includegraphics[width=\linewidth]{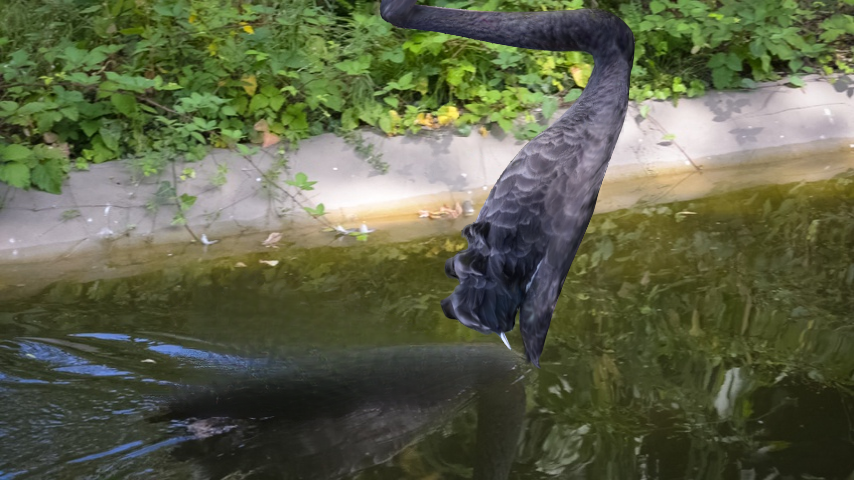}
  \end{subfigure}\hfill
  \begin{subfigure}[b]{0.23\linewidth}
    \includegraphics[width=\linewidth]{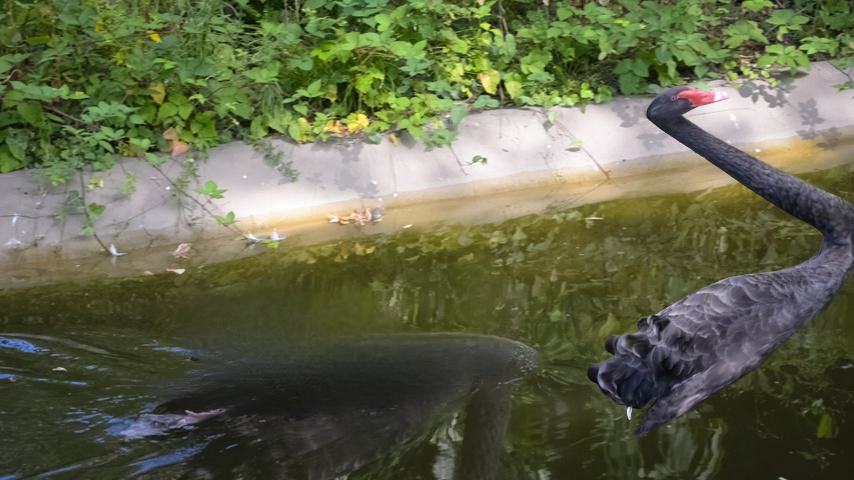}
  \end{subfigure}\hfill
  
    \caption{\textbf{Physics-based editing examples.} The first simulation shows a soft body dropped onto a horizontal surface. The figure shows key moments of the simulation: falling, compression, stretching during bounce-back, and settling. The second simulation shows a spinning soft body bouncing off a horizontal surface.}
    \label{fig:physics_edits}
    \vspace{-0.2cm}
\end{figure}

\paragraph{Temporal Opacity and Our Positioning.}
The most closely related representation is Spacetime
Gaussians~\cite{li2024spacetime}, which also equip each Gaussian with a
temporal opacity, a 1D Gaussian window over time with a learnable center
and scale, so that content can emerge or vanish within a clip. The two
methods, however, are designed for different purposes. Spacetime Gaussians
target real-time novel-view synthesis from calibrated multi-view capture,
and therefore pair their temporal opacity with parametric motion, using
polynomial trajectories for position and rotation so that the geometry
moves through time. \our{} instead targets the editing of monocular,
in-the-wild video, and makes the opposite choice: we introduce no motion
parameters and keep every Gaussian spatially static, modeling all non-rigid
dynamics through temporal opacity alone. This is precisely what makes our
representation editable: a fixed-geometry scene is, by construction, an
ordinary 3DGS asset, so standard editing and simulation tools apply without
modification, a property that motion-parameterized and deformation-based
representations forfeit.

\section{\our{} Video Representation}

Our primary objective is to build a video representation that both accurately reconstructs the input frames and serves as an explicit, editable proxy for the dynamic scene. This is achieved by maintaining static spatial parameters for the 3D Gaussians while representing all scene dynamics exclusively via temporal opacity modulation. Figure~\ref{fig:teaser} provides an overview of \our{}.

\begin{figure}[tb]
    \centering
    \captionsetup[subfigure]{labelformat=empty, skip=1pt, font=small}
    \setlength{\tabcolsep}{0pt}%
    \setlength{\fboxsep}{0pt}\setlength{\fboxrule}{0.05pt}%
    \begin{subfigure}[t]{0.325\linewidth}
        \centering
        \caption{Ground truth}
        \fbox{\includegraphics[width=\linewidth]{}}
    \end{subfigure}\hfill
    \begin{subfigure}[t]{0.325\linewidth}
        \centering
        \caption{VeGaS}
        \fbox{\includegraphics[width=\linewidth]{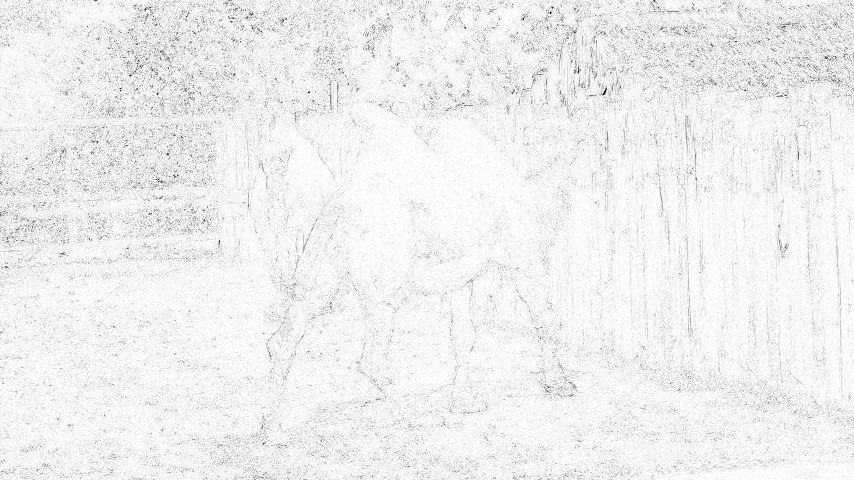}}
    \end{subfigure}\hfill
    \begin{subfigure}[t]{0.325\linewidth}
        \centering
        \caption{\our{}}
        \fbox{\includegraphics[width=\linewidth]{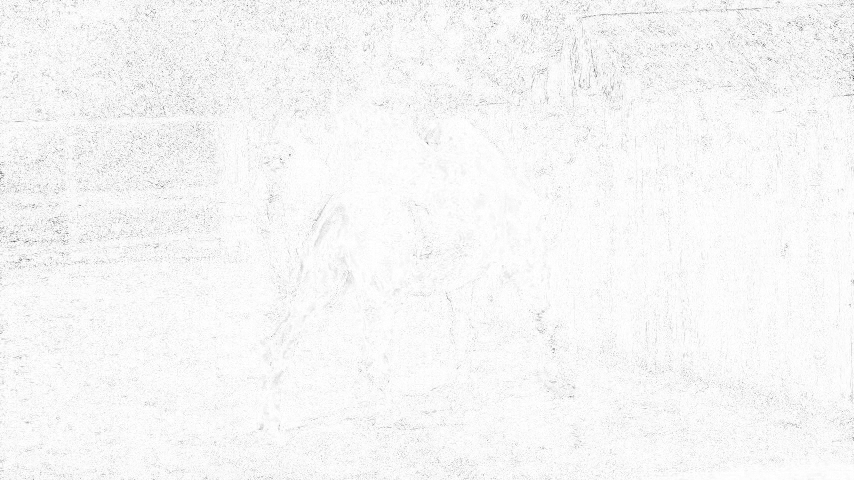}}
    \end{subfigure}
    \caption{\textbf{Frame reconstruction error maps.} Qualitative comparison on a
    frame of the Camel sequence from the DAVIS dataset~\cite{ponttuset20182017davischallengevideo}.
    \emph{Left:} the ground-truth frame. \emph{Center} and \emph{right:} per-pixel absolute error between the ground truth and the reconstructions
    from VeGaS~\cite{smolakdyzewska2026vegas} and \our{}, respectively; darker
    pixels indicate larger error. For clarity, both error maps are scaled by the same
    factor of three. The \our{} error map is markedly fainter than the VeGaS map across the entire frame.}
    \label{fig:diff}
    \vspace{-0.2cm}
\end{figure}

\begin{table*}[!t]
    \centering
    \caption{\textbf{Comparison of \our{} with various methods on the DAVIS~\cite{ponttuset20182017davischallengevideo} validation videos.} Most results are taken from \cite{bond2025gaussianvideo}. The \colorbox{tablered}{best}, \colorbox{orange}{second-best}, and \colorbox{yellow}{third-best} results are highlighted.}
    \begin{tabular}{lccc}
    \toprule
         \textbf{Model} & \textbf{PSNR} $\uparrow$ & \textbf{SSIM} $\uparrow$ & \textbf{LPIPS} $\downarrow$ \\
        \midrule
        \our{} (our) & \best{\textbf{39.80}} & \best{\textbf{0.98}} & \sbest{0.030} \\ 
        {GaussianVideo} \cite{bond2025gaussianvideo} & \sbest{37.38} & \sbest{0.96} & \best{\textbf{0.021}} \\ 
        {GaussianImage} \cite{zhang2024gaussianimage} & \tbest{36.25} & \tbest{0.94} & \tbest{0.045} \\
        {VeGaS} \cite{smolakdyzewska2026vegas} & 34.71 & \tbest{0.94} & 0.084 \\
        {Splatter-a-Video} \cite{sun2024splattervideovideogaussian}& 28.63 & 0.84 & 0.228\\ 
        {NeRV} \cite{chen2021nerv} & 26.15 & 0.72 & 0.312 \\
        {HNeRV} \cite{chen2023hnerv}& 27.82 & 0.72 & 0.252 \\
        {4DGS} \cite{wu20244d} & 18.12 & 0.57 & 0.394 \\ 
        {RoDyF} \cite{liu2023robust} & 24.79 & 0.72 & 0.394 \\ 
        {Deformable Sprites} \cite{ye2022deformable} & 22.83 & 0.70 & 0.301 \\ 
        {OmniMotion} \cite{wang2023tracking} & 24.11 & 0.72 & 0.371 \\ 
        {CoDeF} \cite{ouyang2024codefcontentdeformationfields} & 26.17 & 0.82 & 0.290 \\ 
        \bottomrule
    \end{tabular}%

    \label{table:vr_davis_val}
\end{table*}

\begin{table*}[t]
\small
\centering
\caption{\textbf{Frame reconstruction.} PSNR of the \our{} model under the evaluation protocol of \cite{sun2024splattervideovideogaussian} on various videos from the DAVIS dataset~\citep{ponttuset20182017davischallengevideo}. Baseline results are taken from \cite{smolakdyzewska2026vegas}. The \colorbox{tablered}{best}, \colorbox{orange}{second-best}, and \colorbox{yellow}{third-best} results are highlighted.}
\begin{tabular}{lcccccccc}
\toprule
\textbf{Model} & \textbf{Bear} & \textbf{Cows} & \textbf{Elephant} & \textbf{Breakdance-Flare} & \textbf{Train} & \textbf{Camel} & \textbf{Kite-surf} & \textbf{Average} \\
\midrule
\our{} (our)  & \best{\textbf{33.95}} & \best{\textbf{30.28}} & \best{\textbf{36.11}} & \best{\textbf{38.30}} & \best{\textbf{37.68}} & \best{\textbf{35.28}}  &  \best{\textbf{43.90}} & \best{\textbf{36.50}} \\
VeGaS \citep{smolakdyzewska2026vegas}  &  \sbest{33.23} & \sbest{30.27} & \sbest{33.28} & \sbest{33.19} & \sbest{32.86} & \sbest{32.23}  &  \sbest{38.23} & \sbest{33.31}\\
Splatter-a-Video \citep{sun2024splattervideovideogaussian} & \tbest{30.17} & \tbest{28.24} & \tbest{29.82} & \tbest{27.18} & \tbest{28.09} & \tbest{27.74} & \tbest{27.82} & \tbest{28.44} \\
CoDeF~\citep{ouyang2024codefcontentdeformationfields} & 29.17 & 28.82 & 30.50 & 25.99 & 26.53 & 26.10 & 27.17 & 27.75 \\
Omnimotion \citep{wang2023tracking} & 22.96 & 23.93 & 26.59 & 24.45 & 22.85 & 23.98 & 23.72 & 24.07\\
\bottomrule
\end{tabular}
\label{table:sota-comparison}
\end{table*}

\begin{table*}
\footnotesize
  \centering
  \caption{\textbf{Frame reconstruction.} PSNR and SSIM of the \our{} model under the evaluation setting of \cite{zhao2023dnerv} on various videos from the DAVIS dataset~\cite{ponttuset20182017davischallengevideo}. Baseline results are taken from \cite{smolakdyzewska2026vegas}. The \colorbox{tablered}{best}, \colorbox{orange}{second-}best, and \colorbox{yellow}{third-best} results are highlighted.}
  \begin{tabular}{lcccccccccccc}
 \toprule
       & \multicolumn{2}{c}{NeRV \cite{chen2021nerv}} & \multicolumn{2}{c}{E-NeRV \cite{li2022nerv} } & \multicolumn{2}{c}{HNeRV \cite{chen2023hnerv} } & \multicolumn{2}{c}{DNeRV \cite{zhao2023dnerv}} & \multicolumn{2}{c}{VeGaS \cite{smolakdyzewska2026vegas}} & \multicolumn{2}{c}{ \our{} (our) } \\
       & PSNR$\uparrow$ & SSIM$\uparrow$ & PSNR$\uparrow$ & SSIM$\uparrow$ & PSNR$\uparrow$ & SSIM$\uparrow$ & PSNR$\uparrow$ & SSIM$\uparrow$ & PSNR$\uparrow$ & SSIM$\uparrow$ & PSNR$\uparrow$ & SSIM$\uparrow$\\
    \midrule
    Blackswan  & 28.48 & 0.812 & 29.38 & 0.867 & 30.35 & 0.891 & \tbest{30.92} & \tbest{0.913} & \sbest{34.92} & \sbest{0.932} & \best{\textbf{38.38}} & \best{\textbf{0.968}} \\
    Bmx-bumps  & 29.42 & 0.864 & 28.90 & 0.851 & 29.98 & 0.872 & \tbest{30.59} & \tbest{0.890} & \sbest{33.01} & \sbest{0.915} & \best{\textbf{36.57}} & \best{\textbf{0.955}} \\
    Bmx-trees  & 26.24 & 0.789 & 27.26 & 0.876 & 28.76 & 0.861 & \tbest{29.63} & \tbest{0.882} & \sbest{31.78} & \sbest{0.896} & \best{\textbf{37.28}} & \best{\textbf{0.966}} \\
    Breakdance & 26.45 & 0.915 & 28.33 & 0.941 & 30.45 & \sbest{0.961} & \tbest{30.88} & \best{\textbf{0.968}} & \best{\textbf{32.27}} & 0.950 & \sbest{31.30} & \tbest{0.951} \\
    Camel      & 24.81 & 0.781 & 25.85 & 0.844 & 26.71 & 0.844 & \tbest{27.38} & \sbest{0.887} & \sbest{31.12} & \tbest{0.886} & \best{\textbf{32.61}} & \best{\textbf{0.925}} \\
    Car-roundabout & 24.68 & 0.857 & 26.01 & 0.912 & 27.75 & 0.912 & \tbest{29.35} & \tbest{0.937} & \sbest{32.75} & \sbest{0.941} & \best{\textbf{35.16}} & \best{\textbf{0.962}} \\
    Car-shadow & 26.41 & 0.871 & 30.41 & 0.922 & 31.32 & 0.936 & \tbest{31.95} & \tbest{0.944} & \best{\textbf{36.41}} & \best{\textbf{0.956}} & \sbest{36.10} & \sbest{0.947} \\
    Car-turn   & 27.45 & 0.813 & 29.02 & \tbest{0.888} & 29.65 & 0.879 & \tbest{30.25} & \sbest{0.892} & \sbest{31.44} & 0.852 & \best{\textbf{35.03}} & \best{\textbf{0.924}} \\
    Cows       & 22.55 & 0.702 & 23.74 & 0.819 & 24.11 & 0.792 & \tbest{24.88} & \tbest{0.827} & \sbest{27.97} & \sbest{0.834} & \best{\textbf{28.33}} & \best{\textbf{0.878}} \\
    Dance-twirl& 25.79 & 0.797 & 27.07 & \tbest{0.864} & 28.19 & 0.845 & \tbest{29.13} & \sbest{0.870} & \sbest{30.45} & 0.850 & \best{\textbf{35.08}} & \best{\textbf{0.938}} \\
    Dog        & 28.17 & 0.795 & 30.40 & 0.882 & 30.96 & 0.898 & \tbest{31.32} & \tbest{0.905} & \sbest{34.52} & \sbest{0.914} & \best{\textbf{39.84}} & \best{\textbf{0.977}} \\
  \midrule
    Average    & 26.40 & 0.818 & 27.85 & 0.879 & 28.93 & 0.881 & \tbest{29.66} & \tbest{0.901} & \sbest{32.42} & \sbest{0.902} & \best{\textbf{35.06}} & \best{\textbf{0.945}} \\
   \bottomrule
  \end{tabular}

  \label{table:vr_davis}
\end{table*}

\begin{table*}[t]
\small
\centering
\caption{\textbf{Training statistics and rendering quality metrics for the results in}~Table~\ref{table:sota-comparison}. All experiments were run on a single NVIDIA A100 40GB GPU.}
\begin{tabular}{lccccccc}
\toprule
\textbf{Scene} & \textbf{Total Gaussians} & \textbf{Training Time} & \textbf{Rendering FPS} & \textbf{PSNR} $\uparrow$ & \textbf{SSIM} $\uparrow$ & \textbf{LPIPS} $\downarrow$ \\
\midrule
Bear             & 10.00M & 92 min  & 50.57  & 33.95 & 0.962 & 0.055 \\
Cows             & 10.04M & 107 min & 50.91  & 30.28 & 0.932 & 0.094 \\
Elephant         & 8.23M  & 72 min  & 63.09  & 36.11 & 0.959 & 0.054 \\
Breakdance-Flare & 10.00M & 87 min  & 55.75  & 38.30 & 0.975 & 0.026 \\
Train            & 10.07M & 148 min & 46.79  & 37.68 & 0.987 & 0.007 \\
Camel            & 8.42M  & 80 min  & 58.45  & 35.28 & 0.967 & 0.035 \\
Kite-surf        & 2.80M  & 42 min  & 118.25 & 43.90 & 0.987 & 0.012 \\
\midrule
\textbf{Average} & 8.51M  & 90 min  & 63.40  & 36.50 & 0.967 & 0.040 \\
\bottomrule
\end{tabular}

\label{tab:training_stats}
\end{table*}

\begin{table*}[t]
\small
\centering
\caption{\textbf{Ablation studies on the reconstruction results in~Table~\ref{table:sota-comparison}.} Per-scene PSNR and the mean across all scenes. The \colorbox{tablered}{best}, \colorbox{orange}{second-best}, and \colorbox{yellow}{third-best} results are highlighted.}
\begin{tabular}{lcccccccc}
\toprule
\textbf{Setting} & \textbf{Bear} & \textbf{Cows} & \textbf{Elephant} & \textbf{Breakdance-Flare} & \textbf{Train} & \textbf{Camel} & \textbf{Kite-surf} & \textbf{Average} \\
\midrule
\textbf{\our{} (our)} & \best{\textbf{33.95}} & \sbest{30.28} & \best{\textbf{36.11}} & \best{\textbf{38.30}} & \sbest{37.68} & \best{\textbf{35.28}} & \best{\textbf{43.90}} & \best{\textbf{36.50}} \\
Uniform frame sampling & \sbest{33.50} & \best{\textbf{30.39}} & \sbest{34.78} & \sbest{37.65} & \tbest{36.43} & \sbest{34.22} & \sbest{41.68} & \sbest{35.52} \\
No camera poses & \tbest{30.70} & \tbest{27.16} & \tbest{34.13} & \tbest{35.13} & 31.18 & 27.37 & 37.50 & \tbest{31.88} \\
Opacity resets enabled & 29.28 & 27.00 & 30.95 & 30.94 & 34.17 & \tbest{30.88} & \tbest{37.58} & 31.54 \\
Neural opacity modulation & 26.01 & 24.59 & 28.70 & 27.84 & \best{\textbf{37.85}} & 28.08 & 31.25 & 29.19 \\
No temporal opacity (static 3DGS) & 6.35 & 19.76 & 24.44 & 24.59 & 36.26 & 23.61 & 35.54 & 24.36 \\
\bottomrule
\end{tabular}

\label{tab:ablation-psnr}
\end{table*}

\subsection{Preliminaries: 3D Gaussian Splatting}
3D Gaussian Splatting (3DGS) represents a scene with a set of anisotropic 3D Gaussian primitives:
\begin{equation}
    \mathcal{G} = \{(\mu_i, \Sigma_i, \alpha_i, \mathbf{c}_i)\}_{i=1}^{N},
\end{equation}
where $\mu_i \in \mathbb{R}^3$ is the mean, $\Sigma_i$ is a covariance matrix parameterized as $\Sigma_i = R_i S_i S_i^{\top} R_i^{\top}$ to ensure positive semi-definiteness via a rotation matrix $R_i$ and a diagonal scaling matrix $S_i$, $\alpha_i \in (0,1)$ is the opacity, and $\mathbf{c}_i$ denotes the view-dependent color represented by spherical harmonics coefficients.
For a given camera view, each 3D Gaussian is projected onto the 2D image plane. Pixel colors are rendered via front-to-back alpha compositing over depth-sorted visible Gaussians. This yields a fully differentiable rasterization pipeline optimized by combining an $L_{1}$  pixel-level color loss with structural dissimilarity (D-SSIM).

\subsection{Temporal Opacity Modulation}
In standard 3DGS, each Gaussian is assigned a fixed opacity. To model dynamic scenes while maintaining a static spatial geometry, we redefine opacity as a continuous function of time. For the $i$-th Gaussian, we introduce a learnable temporal mean $\mu _{i}^{\tau }$ and a temporal scale $\sigma _{i}^{\tau }$. The effective opacity at time $t$ is thus formulated as:
\begin{equation}
    \label{eq:temporal-opacity}
    \alpha^\text{b}_i \cdot \exp\left(-\frac{(t - \mu^{\tau}_i)^2}{2 (\sigma^{\tau}_i)^2}\right),
\end{equation}
where $\alpha^\text{b}_{i}$ represents the base learnable opacity. This parameterization acts as a temporal window. By optimizing $\mu^{\tau}_i$ and $\sigma^{\tau}_i$, the model decides exactly when a structural component should appear and disappear. A moving object is thus represented by a sequence of Gaussians lighting up along its trajectory. Figure~\ref{fig:marked_gaussian} illustrates this behavior: a single Gaussian (highlighted in red) contributes to the static background over several consecutive frames and gradually fades out, while the object in the foreground moves. Because the underlying primitives remain standard 3D Gaussians, any spatial edit, including scaling, translation, or duplication, automatically preserves the primitive's temporal duration. This property renders the representation inherently editable.

This low-capacity temporal window represents a strength rather than a limitation. As demonstrated in our ablation studies, replacing this formulation with a neural network that predicts opacity from position and time reduces reconstruction quality, lowering the mean Peak Signal-to-Noise Ratio (PSNR) from 36.50 to 29.19 as shown in Table~\ref{tab:ablation-psnr}. Furthermore, such a network removes the explicit per-Gaussian temporal parameters required for downstream editing.

\subsection{Pose Estimation}\label{subsec:pose}
Because the spatial parameters of the Gaussians in \our{} are static, the camera alone must account for all apparent background motion across the video. We therefore recover a camera for every frame with AnyCam~\cite{wimbauer2025anycamlearningrecovercamera}, running it on the input video and using the predicted per-frame camera-to-world matrices $\{P_t\}_{t=1}^{n}$, with $P_t \in SE(3)$. We adopt these matrices directly as our cameras and hold them fixed throughout training. Fixing the cameras grounds every Gaussian in a single, shared coordinate space, which in turn enables editing: a transformation applied to a Gaussian remains consistent across all frames. Our ablations confirm the importance of this step: replacing the estimated poses with identity matrices reduces the mean PSNR from $36.50$ to $31.88$ (Table~\ref{tab:ablation-psnr}).

\subsection{Optimization Strategy}
We make three key modifications to the standard 3DGS optimization pipeline.

\paragraph{Inverse PSNR Frame Sampling.}
Instead of sampling frames uniformly during training, we track a running PSNR per frame and sample each frame with probability proportional to its inverse PSNR. This adaptive sampling drives the optimization to allocate Gaussians to the most dynamic or complex temporal segments.

\paragraph{Disabling Opacity Resets.}
Standard 3DGS periodically resets opacities to near zero to cull floaters. In our formulation, however, opacity is coupled to the temporal scale $\sigma^{\tau}_i$, so resetting $\alpha^\text{b}_{i}$ destroys the learned temporal windows. We therefore disable the opacity-reset heuristic, which stabilizes training and ensures the convergence of our temporal representation.

\paragraph{Densification Cap.}
To keep memory usage and throughput stable on a single GPU, we halt densification once the model reaches 10M Gaussians. Lifting this cap yields further gains in reconstruction quality, at the cost of higher memory and runtime.

\section{Experiments}

\subsection{Reconstruction Results}

Before demonstrating the editing capabilities of the \our{} representation, we establish its baseline fidelity by evaluating frame reconstruction on the DAVIS dataset~\citep{ponttuset20182017davischallengevideo}. We train our model for 80,000 iterations, with densification continuing until 60,000 iterations, using inverse PSNR sampling and disabled opacity resets.

Tables~\ref{table:vr_davis_val}, \ref{table:sota-comparison}, and \ref{table:vr_davis} present a quantitative comparison against recent video representation methods, including those that support editing. \our{} consistently achieves the highest PSNR and SSIM among all baselines. We also compare against VeGaS~\cite{smolakdyzewska2026vegas} qualitatively. Figure~\ref{fig:diff} shows the per-pixel error maps for a frame of the Camel sequence, alongside the ground truth, for both \our{} and VeGaS. The \our{} map is markedly fainter across the frame, and the object outlines in particular are visibly less pronounced than in the VeGaS map.

Table~\ref{tab:training_stats} provides detailed training statistics. Our model maintains a moderate number of Gaussians (ranging from 2.80M to 10.07M),
while delivering high reconstruction quality across all metrics.

\subsection{Ablation Studies}

We ablate the core modeling and optimization decisions of \our{} while keeping the same training parameters and evaluation protocol as in our main results. Table~\ref{tab:ablation-psnr} reports the final PSNR.

We test the temporal opacity formulation by replacing the per-Gaussian temporal window with a compact learned opacity predictor conditioned on 3D position and time, and by removing temporal opacity altogether (static 3DGS). We also ablate the camera poses by ignoring the estimated trajectory (identity extrinsics), and two training heuristics: switching from inverse-PSNR sampling to uniform sampling and re-enabling periodic opacity resets.

The results show that the temporal opacity mechanism is essential: removing it collapses reconstruction quality (mean PSNR 24.36), while the learned opacity predictor underperforms the temporal opacity modulation (29.19). Camera poses contribute substantially to stability and fidelity (mean PSNR 31.88 without poses). Inverse-PSNR sampling yields consistent gains over uniform sampling (36.50 vs.\ 35.52), and disabling opacity resets remains important for preserving temporal windows (31.54 when resets are enabled).

\subsection{Effect of the Gaussian Budget}
\label{sec:gaussian-budget}

\definecolor{bearcolor}{RGB}{31,119,180}
\definecolor{camelcolor}{RGB}{214,39,40}

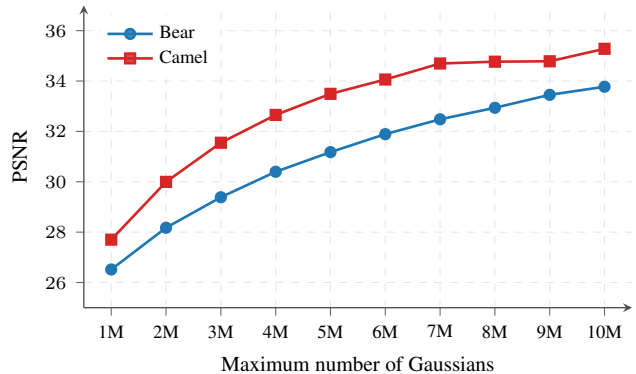
\begin{figure}[t]
  \centering
  \begin{tikzpicture}
    \begin{axis}[
        width=0.87\linewidth, height=4cm,
        scale only axis,
        xmin=0.5, xmax=10.5,
        xtick={1,2,...,10},
        xticklabels={1M,2M,3M,4M,5M,6M,7M,8M,9M,10M},
        xticklabel style={font=\scriptsize},
        ymin=25, ymax=37,
        yticklabel style={font=\scriptsize},
        xlabel={Maximum number of Gaussians},
        ylabel={PSNR},
        xlabel style={font=\footnotesize},
        ylabel style={font=\footnotesize},
        ytick={26,28,30,32,34,36},
        axis x line=bottom, axis y line=left,
        axis line style={black!70, line width=0.5pt},
        tick align=outside,
        major tick length=3pt, minor tick length=1.5pt,
        tick style={black!60, line width=0.4pt},
        xmajorgrids=true, ymajorgrids=true,
        major grid style={black!10, line width=0.35pt, dashed},
        enlarge x limits=false,
        legend cell align=left,
        legend style={
          font=\scriptsize, draw=none, fill=white, fill opacity=0.85,
          text opacity=1, inner sep=2pt, row sep=-1pt,
          at={(0.03,0.97)}, anchor=north west,
        },
      ]
      \addplot[
          color=bearcolor, mark=*, mark size=1.8pt,
          line width=1.0pt, mark options={fill=bearcolor},
        ] coordinates {
          (1,26.5175) (2,28.1751) (3,29.3854) (4,30.3973) (5,31.1771) (6,31.8907) (7,32.4803) (8,32.9385) (9,33.4520) (10,33.7729)
        };
      \addlegendentry{Bear}
      \addplot[
          color=camelcolor, mark=square*, mark size=1.8pt,
          line width=1.0pt, mark options={fill=camelcolor},
        ] coordinates {
          (1,27.7021) (2,29.9960) (3,31.5495) (4,32.6538) (5,33.4914) (6,34.0595) (7,34.6964) (8,34.7674) (9,34.7842) (10,35.2796)
        };
      \addlegendentry{Camel}
    \end{axis}
  \end{tikzpicture}
  \caption{\textbf{Reconstruction quality scales with the Gaussian budget.}
  PSNR as a function of the maximum number of Gaussians, for two DAVIS
  sequences (Bear and Camel); the 10M point is our default cap. Increasing the
  budget consistently improves reconstruction quality.}
  \vspace{-0.2cm}
  \label{fig:max_gaussians_psnr}
\end{figure}

\definecolor{tomfg}{RGB}{202,0,32}
\definecolor{tombg}{RGB}{5,113,176}

\pgfplotsset{
  tomhist/.style={
    width=0.27\linewidth, height=3.7cm,
    scale only axis,
    xmode=log, log basis x=10,
    xmin=5e-4, xmax=5,
    xtick={1e-3,1e-2,1e-1,1e0}, minor x tick num=9,
    ymin=0, ymax=0.32, ytick={0,0.1,0.2,0.3},
    axis x line=bottom, axis y line=left,
    axis line style={black!70, line width=0.5pt},
    tick align=outside,
    major tick length=3pt, minor tick length=1.5pt,
    tick style={black!60, line width=0.4pt},
    xticklabel style={font=\scriptsize},
    yticklabel style={font=\scriptsize,
      /pgf/number format/fixed, /pgf/number format/precision=1},
    title style={font=\small, yshift=-2pt},
    xlabel style={font=\footnotesize},
    ylabel style={font=\footnotesize},
    ymajorgrids=true,
    major grid style={black!10, line width=0.35pt, dashed},
    enlarge x limits=false, enlarge y limits=false,
    legend cell align=left,
    legend style={
      font=\scriptsize, draw=none, fill=white, fill opacity=0.85, text opacity=1,
      inner sep=2pt, row sep=-1pt,
      at={(0.97,0.97)}, anchor=north east,
    },
    legend image post style={line width=1.8pt},
  },
  tombgplot/.style={const plot, mark=none,
    draw=tombg, line width=0.8pt, fill=tombg, fill opacity=0.20},
  tomfgplot/.style={const plot, mark=none,
    draw=tomfg, line width=0.8pt, fill=tomfg, fill opacity=0.20},
}

\begin{figure*}[t]
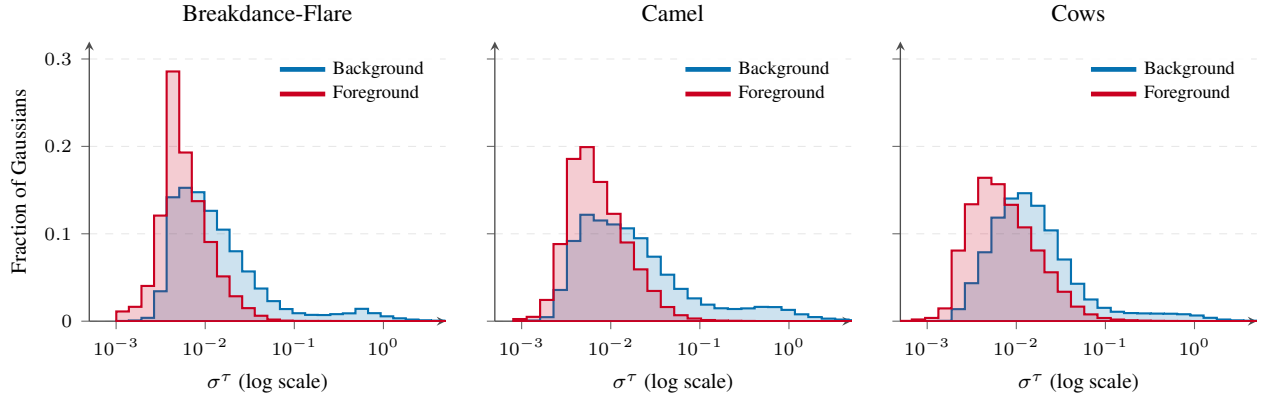

  \centering
  \begin{tikzpicture}
    \begin{groupplot}[
        group style={
          group size=3 by 1, horizontal sep=0.65cm,
          ylabels at=edge left, yticklabels at=edge left,
        },
        tomhist,
        xlabel={$\sigma^{\tau}$ (log scale)},
      ]
      \nextgroupplot[title={Breakdance-Flare}, ylabel={Fraction of Gaussians}]
        \input{figures/time_scale/time_scale_log_breakdance-flare.tex}
        \addlegendentry{Background}
        \addlegendentry{Foreground}
      \nextgroupplot[title={Camel}]
        \input{figures/time_scale/time_scale_log_camel.tex}
        \addlegendentry{Background}
        \addlegendentry{Foreground}
      \nextgroupplot[title={Cows}]
        \input{figures/time_scale/time_scale_log_cows.tex}
        \addlegendentry{Background}
        \addlegendentry{Foreground}
    \end{groupplot}
  \end{tikzpicture}
  \caption{\textbf{Foreground and background Gaussians occupy distinct temporal scales.}
  Count-normalized histograms of the learned temporal scale $\sigma^{\tau}$ for
  models trained on the foreground (\textcolor{tomfg}{red}) vs.\ the background
  (\textcolor{tombg}{blue}) across three DAVIS sequences: Breakdance-Flare, Camel, and Cows. In all three the two populations
  are clearly separated: foreground Gaussians cluster at small $\sigma^{\tau}$, switching on
  briefly to follow the moving object, whereas background Gaussians shift toward larger values,
  consistent with covering a mostly static region.}
  \label{fig:time_scale_log}
\end{figure*}

\newcommand{\cropimg}[2][]{%
    \includegraphics[
        width=\linewidth,
        trim=65mm 11mm 95mm 33mm, %
        clip,
        #1
    ]{#2}
}

\begin{figure*}[t]
    \centering
    \setlength{\tabcolsep}{2pt}
    \begin{subfigure}{0.24\textwidth}
        \cropimg{}\\[-2mm]
        \caption*{$t$}
    \end{subfigure}\hspace{2pt}
    \begin{subfigure}{0.24\textwidth}
        \cropimg{}\\[-2mm]
        \caption*{$t+2$}
    \end{subfigure}\hspace{2pt}
    \begin{subfigure}{0.24\textwidth}
        \cropimg{}\\[-2mm]
        \caption*{$t+4$}
    \end{subfigure}\hspace{2pt}
    \begin{subfigure}{0.24\textwidth}
        \cropimg{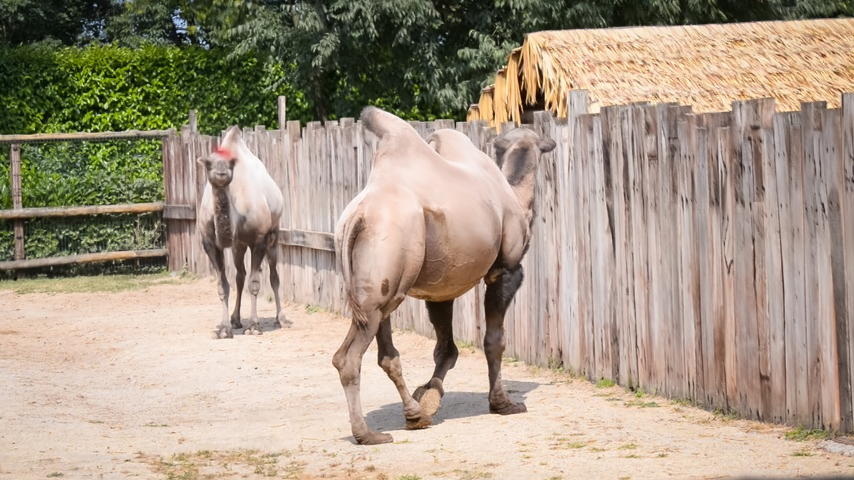}\\[-2mm]
        \caption*{$t+6$}
    \end{subfigure}
    \caption{\textbf{A single Gaussian contributes to multiple frames.} We show every other frame from a run of seven consecutive frames, with one Gaussian highlighted in red (top-left corner). The Gaussian gradually fades in and out, contributing to the mostly static background over several frames.}
    \label{fig:marked_gaussian}
    \vspace{-0.2cm}
\end{figure*}

We study how the maximum number of Gaussians affects reconstruction quality by training \our{} with caps ranging from 1M to 10M Gaussians, keeping all other hyperparameters fixed. Figure~\ref{fig:max_gaussians_psnr} shows PSNR as a function of the Gaussian budget for the Bear and Camel sequences from DAVIS. Quality improves consistently as more Gaussians become available, with the steepest gains in the low-budget regime, where the model must aggressively prioritize which temporal windows to represent.

\subsection{Temporal Scale Separation}
\label{sec:temporal-separation}

A central premise of \our{} is that temporal opacity modulation lets a single static geometry encode motion: foreground Gaussians, which track the moving object, should remain opaque only briefly, whereas background Gaussians, covering comparatively static content, can persist over much longer spans. We examine this behavior by inspecting the learned temporal scale $\sigma^{\tau}$, which directly controls how long each Gaussian stays visible (Eq.~\ref{eq:temporal-opacity}).

To isolate the foreground and background, we train two variants per scene that differ only in which pixels supervise the reconstruction: a \emph{foreground} variant, supervised on the moving object, and a \emph{background} variant, supervised on the remaining, mostly static, region.

Figure~\ref{fig:time_scale_log} shows the resulting $\sigma^{\tau}$ distributions across three DAVIS scenes. In all cases the foreground and background populations are visibly separated on the log axis: foreground Gaussians concentrate at small $\sigma^{\tau}$, switching on only briefly to follow the moving object, while background Gaussians shift toward larger values, reflecting the mostly static nature of the backdrop. This emergent separation arises without any explicit supervision on $\sigma^{\tau}$, demonstrating that temporal opacity modulation naturally encodes the distinction between dynamic foreground content and its surroundings.

\section{Video Editing}

\begin{figure}
      \centering
  \captionsetup[subfigure]{labelformat=empty, skip=1pt}

  \makebox[14pt]{\rotatebox{90}{\scriptsize Original}}%
  \begin{subfigure}[b]{0.30\linewidth}
    \includegraphics[width=\linewidth]{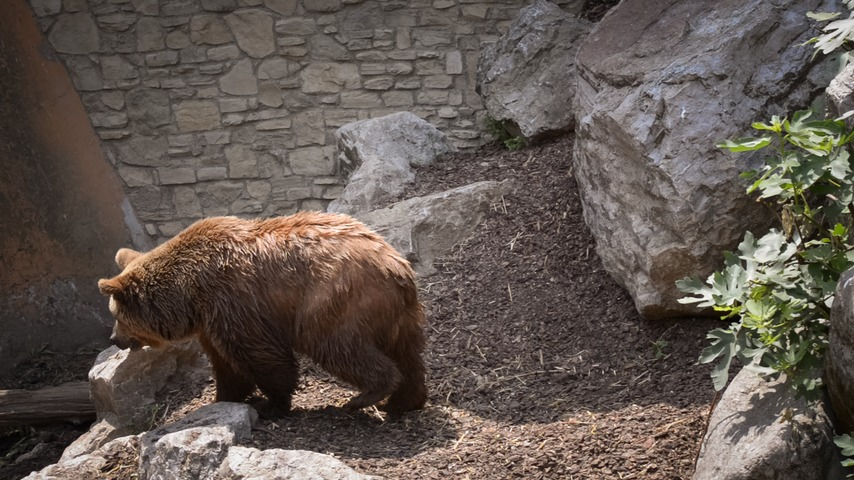}
  \end{subfigure}\hfill
  \begin{subfigure}[b]{0.30\linewidth}
    \includegraphics[width=\linewidth]{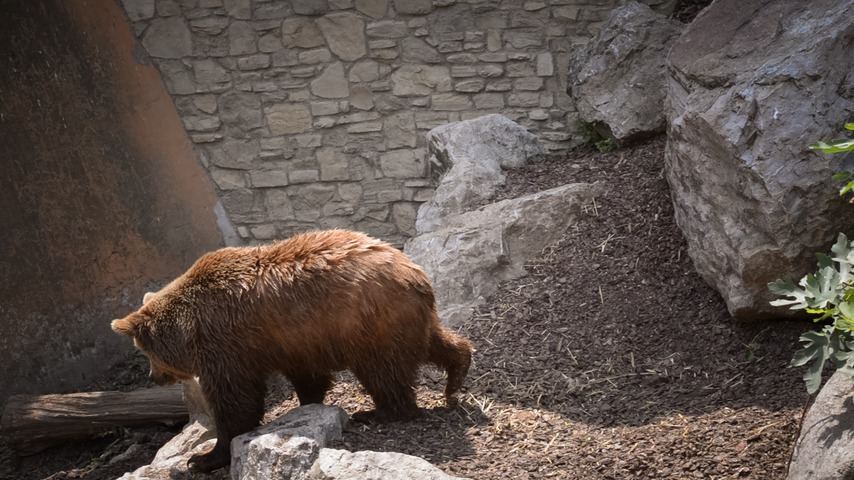}
  \end{subfigure}\hfill
  \begin{subfigure}[b]{0.30\linewidth}
    \includegraphics[width=\linewidth]{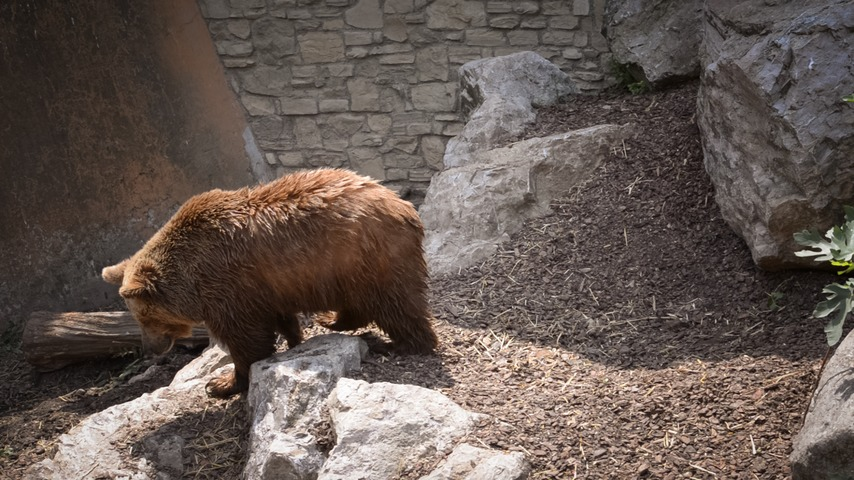}
  \end{subfigure}

  \makebox[14pt]{\rotatebox{90}{\scriptsize Scale}}%
  \begin{subfigure}[b]{0.30\linewidth}
    \includegraphics[width=\linewidth]{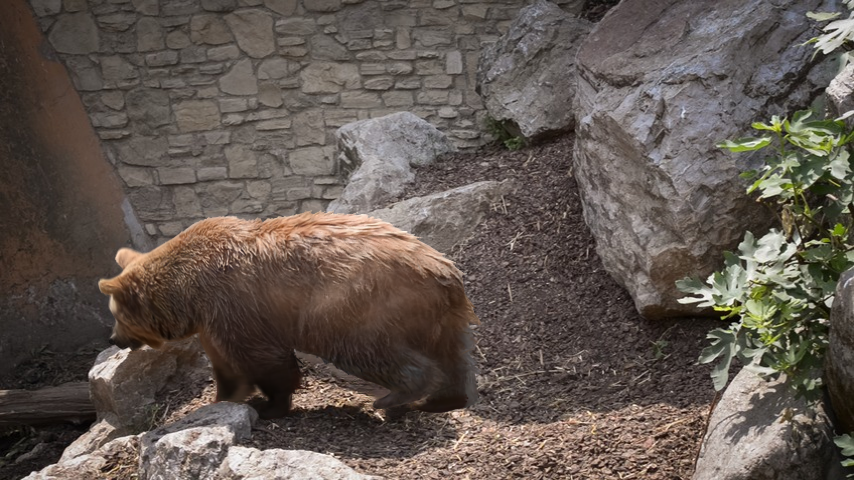}
  \end{subfigure}\hfill
  \begin{subfigure}[b]{0.30\linewidth}
    \includegraphics[width=\linewidth]{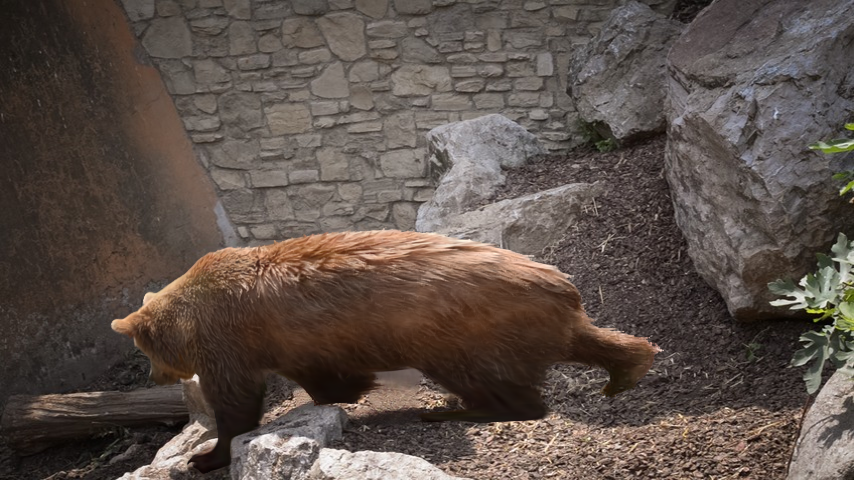}
  \end{subfigure}\hfill
  \begin{subfigure}[b]{0.30\linewidth}
    \includegraphics[width=\linewidth]{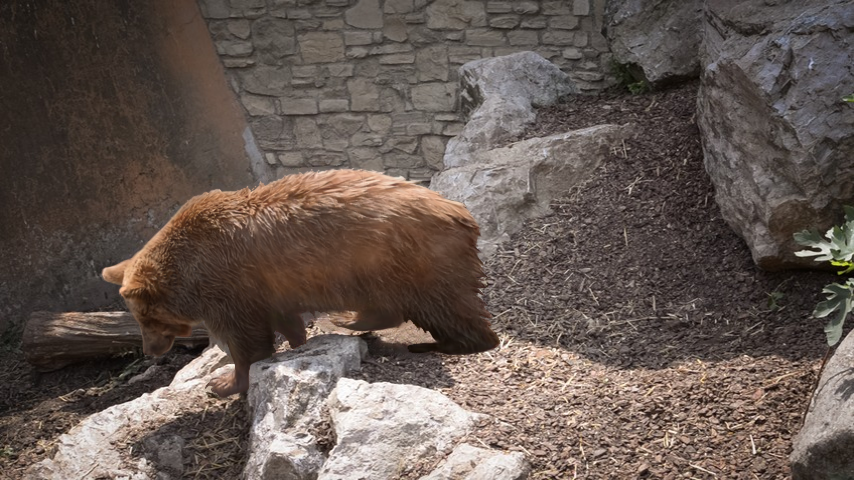}
  \end{subfigure}

  \vspace{1pt}

  \makebox[14pt]{\rotatebox{90}{\scriptsize Original}}%
  \begin{subfigure}[b]{0.30\linewidth}
    \includegraphics[width=\linewidth]{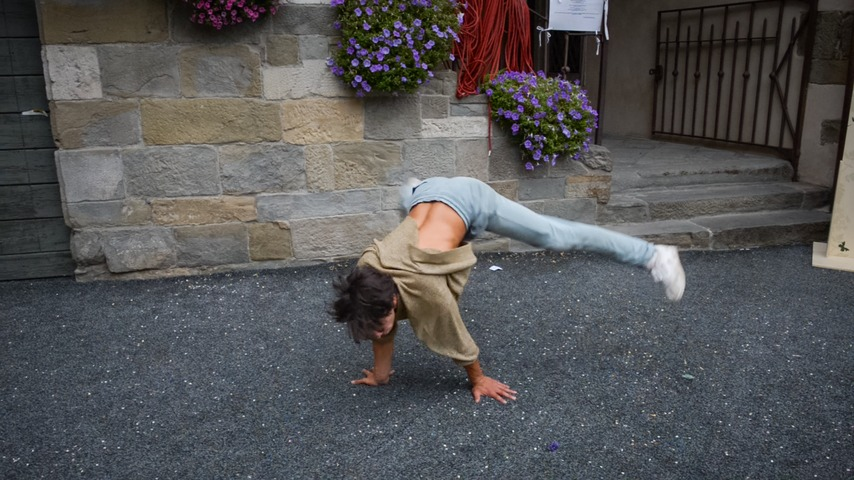}
  \end{subfigure}\hfill
  \begin{subfigure}[b]{0.30\linewidth}
    \includegraphics[width=\linewidth]{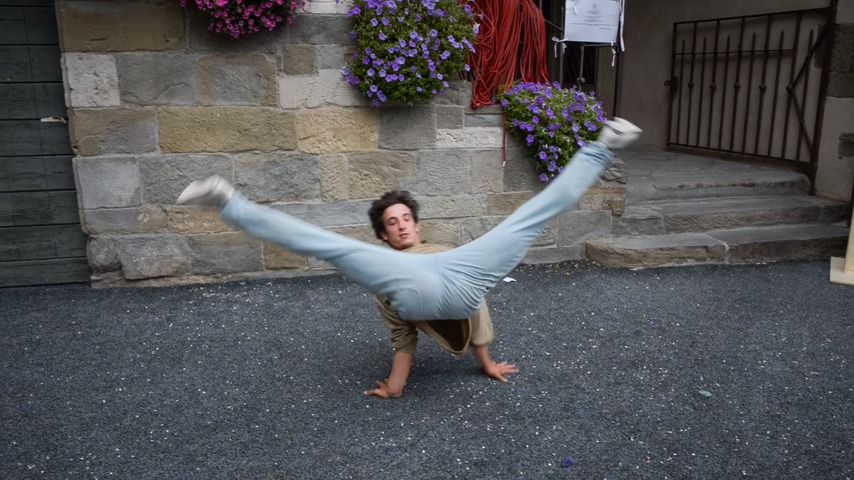}
  \end{subfigure}\hfill
  \begin{subfigure}[b]{0.30\linewidth}
    \includegraphics[width=\linewidth]{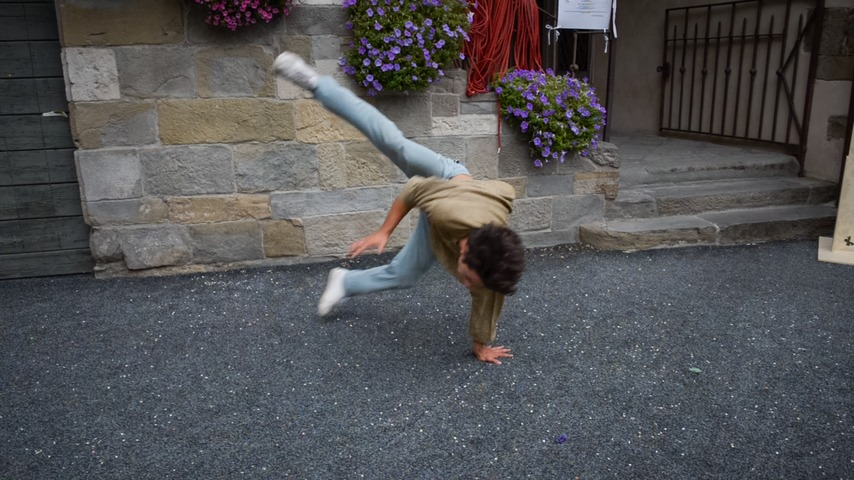}
  \end{subfigure}

  \makebox[14pt]{\rotatebox{90}{\scriptsize Multiply}}%
  \begin{subfigure}[b]{0.30\linewidth}
    \includegraphics[width=\linewidth]{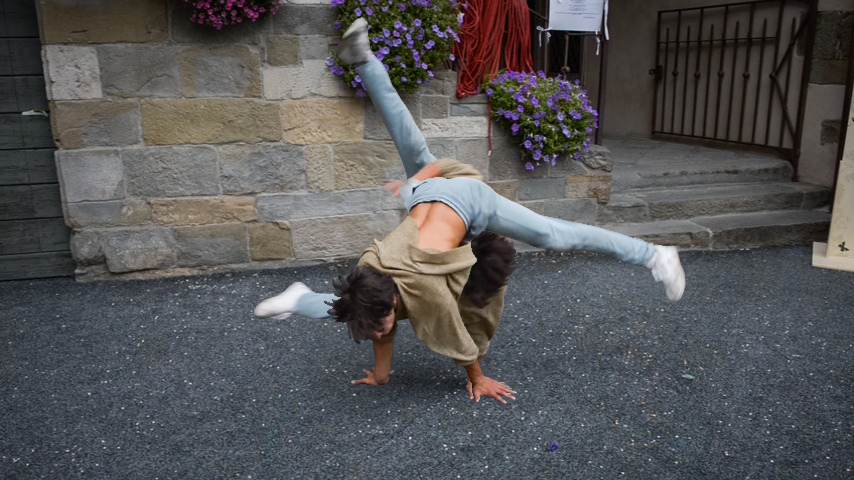}
  \end{subfigure}\hfill
  \begin{subfigure}[b]{0.30\linewidth}
    \includegraphics[width=\linewidth]{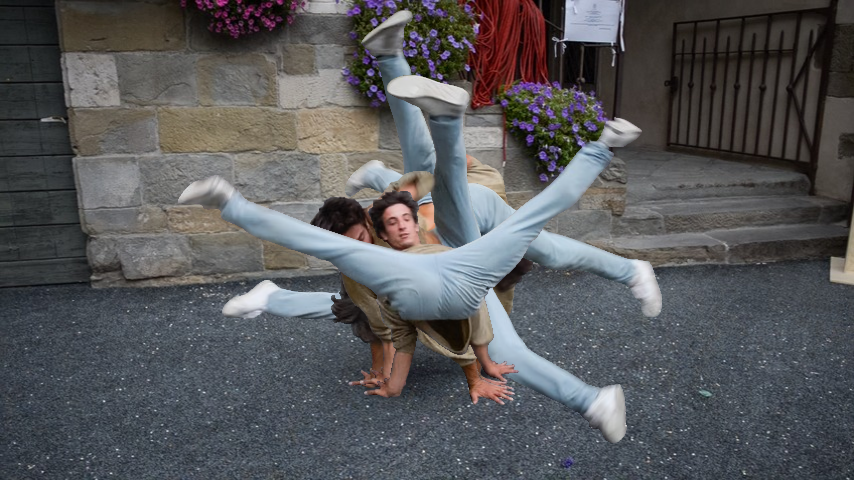}
  \end{subfigure}\hfill
  \begin{subfigure}[b]{0.30\linewidth}
    \includegraphics[width=\linewidth]{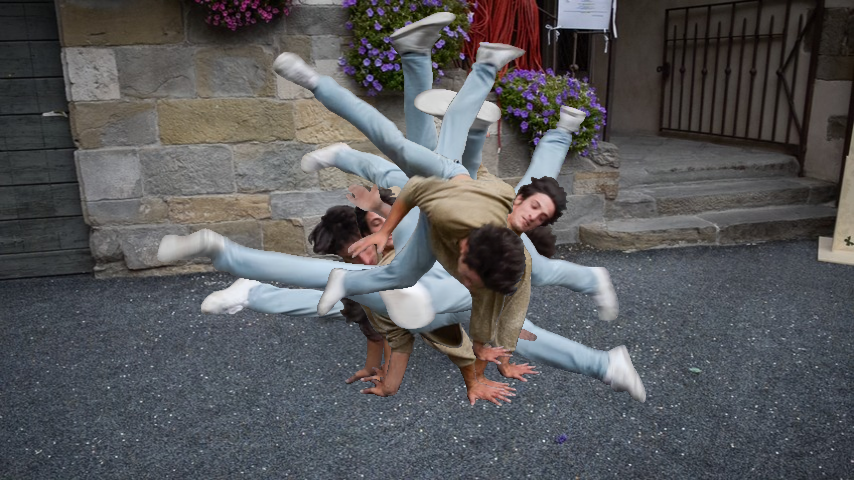}
  \end{subfigure}

  \vspace{1pt}

  \makebox[14pt]{\rotatebox{90}{\scriptsize Original}}%
  \begin{subfigure}[b]{0.30\linewidth}
    \includegraphics[width=\linewidth]{}
  \end{subfigure}\hfill
  \begin{subfigure}[b]{0.30\linewidth}
    \includegraphics[width=\linewidth]{}
  \end{subfigure}\hfill
  \begin{subfigure}[b]{0.30\linewidth}
    \includegraphics[width=\linewidth]{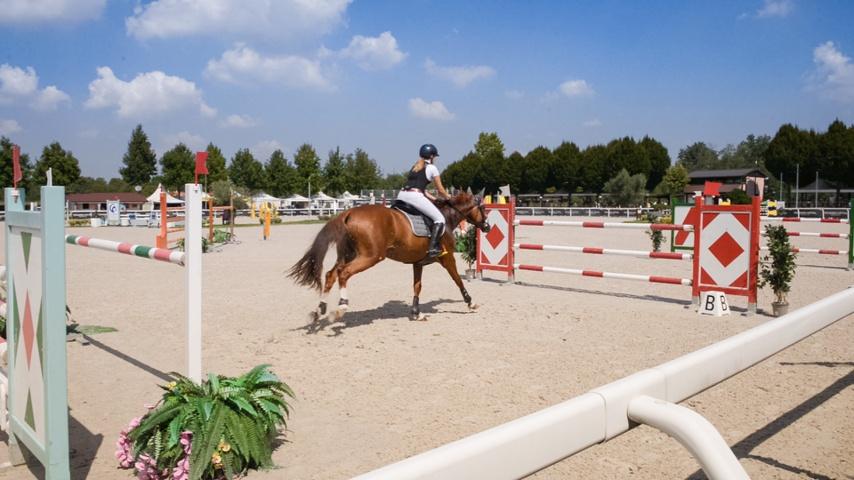}
  \end{subfigure}

  \makebox[14pt]{\rotatebox{90}{\scriptsize Vanishing}}%
  \begin{subfigure}[b]{0.30\linewidth}
    \includegraphics[width=\linewidth]{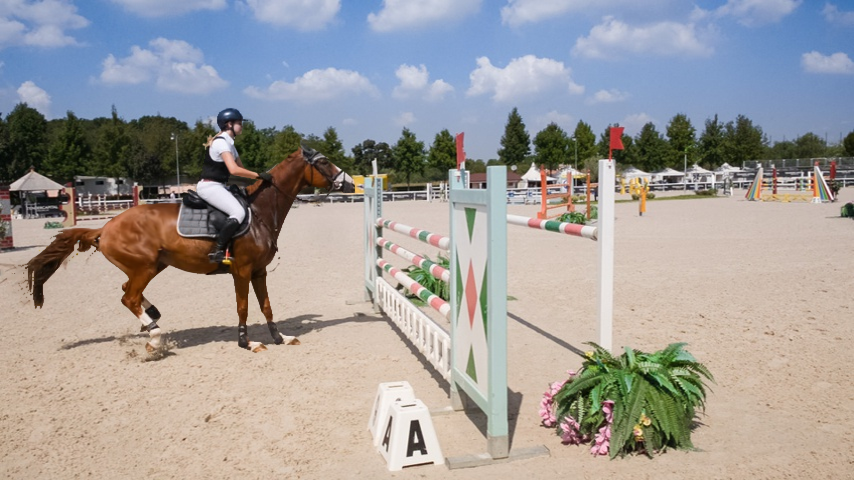}
  \end{subfigure}\hfill
  \begin{subfigure}[b]{0.30\linewidth}
    \includegraphics[width=\linewidth]{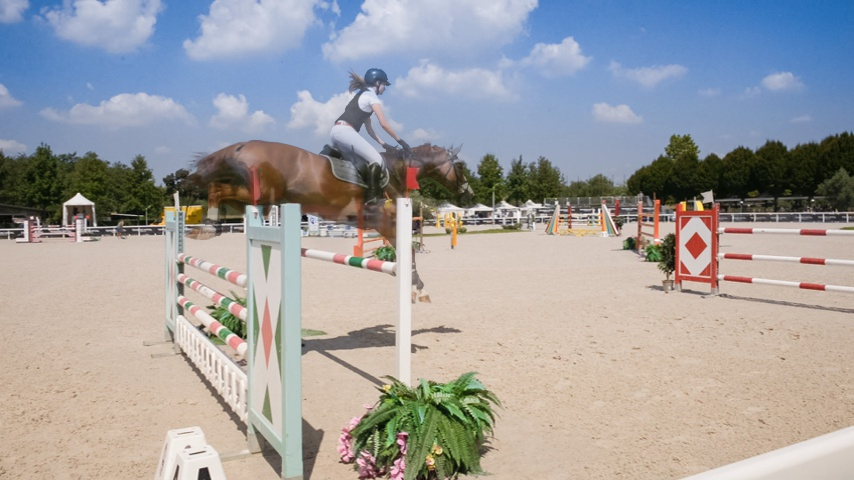}
  \end{subfigure}\hfill
  \begin{subfigure}[b]{0.30\linewidth}
    \includegraphics[width=\linewidth]{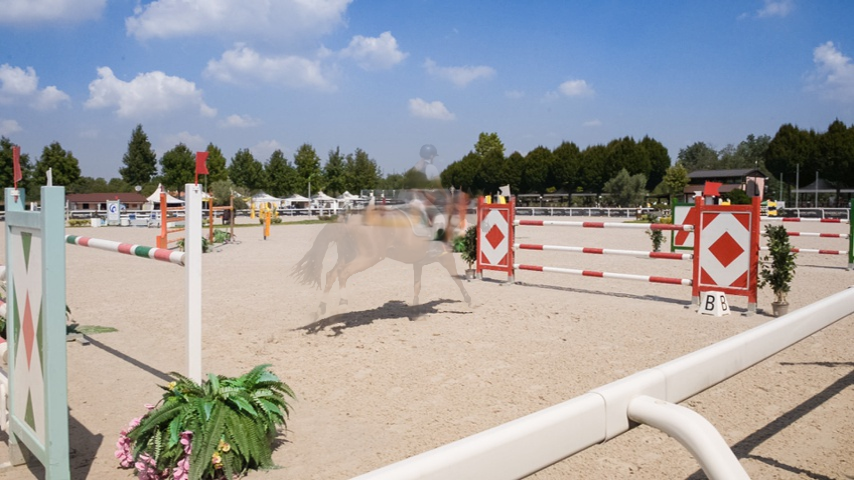}
  \end{subfigure}
    \caption{\textbf{Manual video editing examples.} \our{} supports editing of dynamic scenes: scaling (top), duplication (middle), and removal (bottom), each shown below its original sequence.}
    \label{fig:edits}
    \vspace{-0.2cm}
\end{figure}

Because the \our{} representation is built on standard 3D spatial Gaussians, it
naturally admits explicit geometric modification. In this section we describe
how the explicit temporal formulation of our Gaussians enables intuitive
editing of a dynamic scene over time, including operations such as object
duplication, scaling, and removal, all without costly re-rendering or traversal
of a latent space (Figure~\ref{fig:edits}).

\input{figures/edit_pipeline}

We first apply video segmentation masks to separate the foreground object of
interest from the background. Then, our model is trained on foreground objects under
additional constraints. The training follows a setup similar to that of MiRaGe~\cite{waczynska2025mirage}: a dual-camera arrangement supervises a cloud of Gaussians constrained
to a plane lying between the two cameras. Although \our{} incorporates estimated
camera poses, we omit them here, as they are computed over the entire frame and
thus model the motion of the background rather than the foreground object.
Restricting the Gaussians to a single plane yields a simple and convenient
editing scheme.

The foreground Gaussian cloud is first converted into a set of tetrahedra,
extending the flat-Gaussian-to-triangle conversion demonstrated
in GaMeS~\cite{waczynska2024games}. These tetrahedra encode the geometric parameters of the
Gaussians, namely their means $\mu_i$ and covariances $\Sigma_i$. This
representation simplifies editing considerably: only the tetrahedra vertices
need to be displaced, and the position, rotation, and scale of each Gaussian
follow directly from that displacement.

Once an edit is applied, the tetrahedra are converted back into Gaussians. The
geometric parameters $\mu_i, \Sigma_i$ are read from the modified tetrahedra,
while the remaining parameters $\alpha^\text{b}_i$, $\mathbf{c}_i$, $\mu^\tau_i$, and
$\sigma^\tau_i$ are copied from the corresponding unedited Gaussians. The result
is an edited Gaussian cloud, which can be rendered directly. When the edit varies over time, this procedure is applied per frame, yielding a per-frame
Gaussian cloud. Figure~\ref{fig:edit_pipeline} provides an overview of the editing pipeline.

\subsection{Physics-Based Edits}

Because the tetrahedral representation decouples the spatial geometry from the
temporal formulation, the same structure that enables manual editing can also be
driven by a physics simulation, bridging the gap between video and physics
engines. Object motion is then governed by physical laws rather than by
hand-specified transformations.

We export the tetrahedra as a Stanford PLY point cloud and import them
into external software with a physics engine, such as Blender\footnote{\url{https://www.blender.org}}. The
simulation acts on the vertex positions, producing a sequence of displaced
tetrahedra, one per simulated frame. These are converted back into Gaussians
using the same procedure as for manual edits, and the resulting per-frame clouds
are rendered to obtain the final sequence.
Figure~\ref{fig:physics_edits} shows physics-based editing examples.

\section{Conclusion}
We presented \our, an approach for creating editable video representations from unconstrained videos. By equipping standard 3D Gaussians with a learnable temporal opacity formulation and stabilizing the scene through robust AnyCam pose estimation, our method circumvents the need for complex deformation fields or folded distributions. The resulting representation outperforms existing editable video methods in visual fidelity on the DAVIS dataset and provides an explicit 3D structure that natively supports advanced downstream tasks such as manual structural modifications and physics-based edits.

\paragraph{Limitations}
As with other video models based on Gaussian Splatting \cite{sun2024splattervideovideogaussian, smolakdyzewska2026vegas}, \our{} faces challenges when reconstructing extended video sequences and abrupt, significant changes. Furthermore, the method can require a substantial number of Gaussians, which consequently increases memory consumption and inference time for longer videos.

\section*{Acknowledgments}
The work of K. Howil and P. Spurek was supported by the project \textit{Effective Rendering of 3D Objects Using Gaussian Splatting in an Augmented Reality Environment} (FENG.02.02-IP.05-0114/23), carried out under the First Team programme of the Foundation for Polish Science and co-financed by the European Union through the European Funds for Smart Economy 2021–2027 (FENG). The work of M. Mazur was supported by the National Centre of Science, Poland Grant No. 2021/43/B/ST6/01456. We gratefully acknowledge Polish high-performance computing infrastructure PLGrid (HPC Centers: ACK Cyfronet AGH, CI TASK, WCSS) for providing computer facilities and support within computational grant no. PLG/2026/019272.

{
    \small
    \bibliographystyle{ieeenat_fullname}
    \bibliography{main}
}

\end{document}